# Hidden Structure and Function in the Lexicon

Olivier Picard[1], Mélanie Lord[1,] Alexandre Blondin-Massé[2], Odile Marcotte[3,6], Marcos Lopes[4]
and Stevan Harnad[1,5]

[1] Département de psychologie, Université du Québec à Montréal
`picard.olivier.2@courrier.uqam.ca`
`{lord.melanie, harnad}@uqam.ca`
`http://wwd.crcsc.uqam.ca`
[2] Département de mathématique, Université du Québec à Chicoutimi
`ablondin@uqac.ca`
`http://thales.math.uqam.ca/~blondin/`
[3] Département d'informatique, Université du Québec à Montréal
[4] Department of Linguistics, Universidade de São Paulo, Brazil
`marcoslopes@usp.br`
[5] Department of Electronics and Computer Science, University of Southampton, UK
`harnad@ecs.soton.ac.uk`
[6] Centre de recherches mathématiques, Université de Montréal
`Odile.Marcotte@gerad.ca`

**Abstract**. How many words are needed to define all the words in a dictionary? Graph-theoretic analysis reveals that about 10% of a dictionary is a unique Kernel of words that define one another and all the rest, but this is not the smallest such subset. The Kernel consists of one huge strongly connected component (SCC), about half its size, the Core, surrounded by many small SCCs, the Satellites. Core words can define one another but not the rest of the dictionary. The Kernel also contains many overlapping Minimal Grounding Sets (MGSs), each about the same size as the Core, each part-Core, part-Satellite. MGS words can define all the rest of the dictionary. They are learned earlier, more concrete and more frequent than the rest of the dictionary. Satellite words, not correlated with age or frequency, are less concrete (more abstract) words that are also needed for full lexical power.

## 1 Introduction

Dictionaries catalogue and define the words of a language.[1] In principle, since every word in a dictionary is defined, it should be possible to learn the meaning of any word through verbal definitions alone (Blondin-Massé et al. 2013). However, in order to understand the meaning of the word that is being defined, one has to understand the meaning of the words used to define it. If not, one has to look up the definition of those words too. But if one has to keep looking up the definition of each of the words used to define a word, and then the definition of each of the words that define the words that define the words, and so on, one will eventually come full circle, never having learned a meaning at all.

This is the *symbol grounding problem*: The meanings of all words cannot be learned through definitions alone (Harnad 1990). The meanings of some words, at least, have to be "grounded" by some other means than verbal definitions. That other means is probably direct sensorimotor experience (Harnad 2010), but the learning of categories from sensorimotor experience is not the subject of this paper. Here we ask only *how many words* need to be known by some other means such that all the rest can be learned via definitions composed only of those already known words, and *how do those words differ from the rest*?

## 2 Dictionary Graphs

To answer this question dictionaries can be analyzed using graph theory. These analyses have begun to reveal a hidden structure in dictionaries that was not previously known (see Fig. 1). By recursively removing all the words that are defined but do not define any further word, every dictionary can be reduced by about 90% to a unique set of words (which we have called the *Kernel*) from which all the words in the dictionary can be defined (Blondin-Massé et al. 2008). There is only one such Kernel in any dictionary, but *the Kernel is not the smallest number of words out of which the whole dictionary can be defined*. We call such a smallest subset of words a *Minimal Grounding Set* (MGS). (In graph theory it is called a "minimum feedback vertex set"; Karp 1972; Lapointe et al. 2012.) The MGS is about half the size of the Kernel (Table 2), but, unlike the Kernel, it is not unique: There are a huge number of (overlapping) MGSs in every dictionary, each of the same minimal size; each is a subset of the Kernel and any one of the MGSs grounds the entire dictionary.

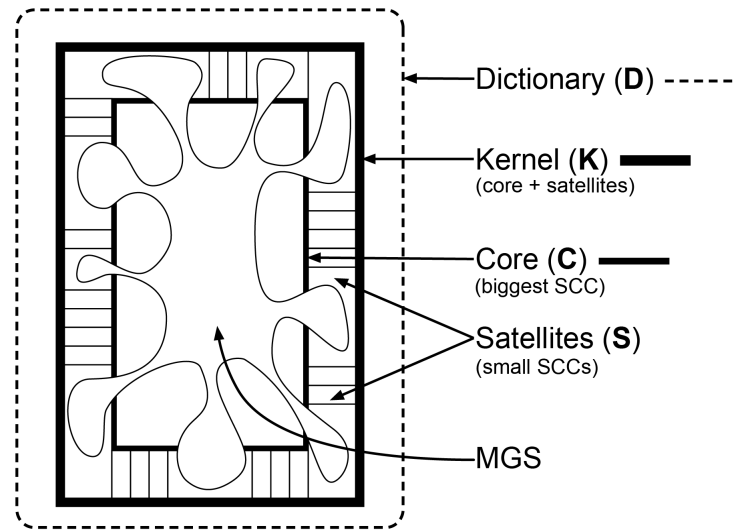

**Fig. 1**. **Diagram of the hidden structure of dictionaries**. Every dictionary (**D**) tested so far (see Table 2) contains a *Kernel* (**K**) of words (fewer than 10% of of the dictionary) from which all the words in the dictionary can be defined. But the Kernel is not the smallest number of words that can define all the rest. In the graph of the Kernel, a directional link means that one word defines another word. The Kernel consists of many subsets in which every word is connected to every other word via a chain of definitional links. Each such subset is called a *Strongly Connected Component* (SCC). About half of the Kernel consists of one big SCC, the *Core* (**C**). The rest of the Kernel is small SCCs (*Satellites*) (**S**) surrounding the Core. The Core alone can define all of its own words, but not all the rest of the words in the Kernel (hence it cannot define the dictionary as a whole either). Solving a graph-theoretic problem for the Kernel of a dictionary (finding its "minimal feedback vertex set") reveals the smallest number of words from which all the rest of its words can be defined: the *Minimal Grounding Set* (MGS). The MGS is also about half the size of the Kernel, but there are a huge number of overlapping MGSs in the Kernel, each of which includes words from both the Core and its Satellites (but only one of the MGSs is illustrated here). The words in these different structural components of the Dictionary Graph turn out to have different psycholinguistic properties (Figs. 2 & 3). (Note that the diagram is not drawn to scale, as **K** is really only 10% of **D**.)

The Kernel, however, is not just a large number of overlapping MGSs. It has structure too. It consists of a large number of strongly connected components (SCCs). (A directed graph -- in which a directional link indicates that word A belongs to the definition of word B -- is "strongly connected" if every word in the graph can be reached by a chain of definitional links from any other word in the graph.) Most of the SCCs of the Dictionary's Kernel are small, but in every dictionary we have analyzed so far there also turns out to be one very large SCC, about half the size of the Kernel. We call this the Kernel's *Core*[2].

The Kernel itself is a self-contained dictionary, just like the dictionary as a whole: every word in the Kernel can be fully defined using only words in the Kernel. The Core is likewise a self-contained dictionary; but the Core is also an SCC (at least in all the full-size dictionaries of natural languages that we have so far examined[3]), whereas the Kernel is not: Every word within the Core can be reached by a chain of definitions from any other word in the Core. In what follows, our statements about the Core will assume that we are discussing full-size dictionaries of natural languages (unless stated otherwise).

The Kernel is a Grounding Set for the dictionary as a whole, but it is not a *Minimal* Grounding Set (MGS) for the dictionary as a whole. The Core, in contrast, is not only *not* an MGS for the dictionary as a whole: it is not even a Grounding Set at all. The words in the Core alone are not enough to define all the rest of the words in the dictionary, outside the Core.

In contrast, the MGSs -- which, like the Core, are each about half the size of the Kernel -- are each contained within the Kernel, but none is completely contained within the Core: Each MGS straddles the Core and the surrounding "Satellite" layer of smaller SCCs. Each MGS can define all the rest of the words in the dictionary, but no MGS is an SCC (see Fig. 1 & Table 1).

The MGSs of Dictionaries hence turn out empirically[4] to consist of words that are partly in the Core (which is entirely within the Kernel) and partly in the remainder of the Kernel (K) -- the part outside the Core (C), which we call the *Satellites* (S) (K minus C) because they consist of much smaller SCCs, encircling the huge Core. The MGSs, the smallest subsets capable of defining all the rest of the dictionary, are hence part-Core and part-Satellite. The natural question, then, is: Is there any difference between the *kinds* of words that are in the various components of this hidden structure of the dictionary: the MGSs, the Core, the Satellites, the Kernel, and the rest of the dictionary outside the Kernel?

Associated with this hidden graph-theoretic structure of the dictionary some evidence of hidden psycholinguistic function is beginning to emerge. It turns out that the words in the Kernel are learned at a significantly younger age, and are more concrete and frequent than the words in the rest of the Dictionary. The same is true, but moreso, comparing the Core with the rest of the dictionary, and still moreso comparing the MGSs with the rest of the dictionary (Fig. 2, left). There are hints, however, that something more subtle is also going on: All five psycholinguistic variables are themselves highly inter-correlated. If we factor out their inter-correlations and look only at their independent effects, the relationships with the hidden structure of the dictionary change subtly: The words in the Kernel remain younger and more frequent than the rest of the dictionary, but once the variance correlated with age is removed, for the residual variance the Kernel is *more abstract* than the rest of the Dictionary. In contrast, this reversal does not happen for either the Core or the MGSs. Hence the locus of the reversal is the Satellite layer (Fig. 2, right.). We will now describe the analyses that generated this pattern of results.

**Table 1.** Practically speaking, a *Dictionary* (D) is a set of words and their definitions in which all the defined and defining words are defined. By recursively removing words that are not used to define further words and that can be reached by definition from the remaining words, a Dictionary can be reduced by about 90% to a *Kernel* (K) of words from which all the other words can be defined. The Kernel is hence a *Grounding Set* (GS) of a Dictionary: a subset that is itself a Dictionary, and that can also define all the words in the rest of the Dictionary. A *Strongly Connected Component* (SCC) of a Dictionary graph is a subset in which there is a definitional path from every word to every other word in the subset. The Kernel's Core (C) is the union of all the strongly connected components (SCCs) of the Kernel that do not receive any incoming definitional links from outside themselves. *Minimal Grounding Sets* (MGSs) are the smallest-sized subsets of words that can define all the words in the rest of the Dictionary. (Any Dictionary has only one Kernel but many MGSs. In all full dictionaries analyzed so far, the Core has always been an SCC, but in some mini-dictionaries generated by the online Dictionary game the Core was not an SCC, but a disjoint union of SCCs).

| a ⇒ is necessarily a ⇓ | **D**ict | **K**ern | **GS** | **SCC** | **C**ore | **MGS** |
|---|---|---|---|---|---|---|
| Dictionary (**D**) | x | x | - | - | x | - |
| Grounding Set (**GS**) | x | x | x | - | - | x |
| Strongly Connected Component (**SCC**) | - | - | - | x | x | - |
| Minimal Grounding Set (**MGS**) | - | - | - | - | - | x |

## 3 Psycholinguistic Properties of Hidden Structures

The MRC database (Wilson 1987) provides psycholinguistic data for words of the English language, including (1) average age of acquisition, (2) degree of concreteness (vs. abstractness), (3) written frequency, (4) oral frequency and (5) degree of (visual) imageability. Analyses of Variance (ANOVAs) reveal that the words in the Kernel (K) differ significantly ($p < .001$) from words in the rest of the Dictionary (D) for all five variables: Kernel words are learned significantly younger, more concrete, more frequent, both orally and in writing, and more imageable than words in the rest of the dictionary. The same effect was found for all five variables in comparing Core (C) words with the rest of the dictionary as well as in comparing MGS words with the rest of the dictionary. The effect was likewise found in comparing Core words with the rest of the Kernel (the Satellites, S) rather than the rest of the dictionary, as well as in comparing MGS words with the rest of the Kernel rather than the rest of the dictionary. Hence the conclusion is that the effects get stronger as one moves from Dictionary to Kernel to Core to MGS for each of the five psycholinguistic variables, as schematized on the left part of Fig. 2. (Three different MGSs were tested, with much the same result.)

One can summarize these findings as the relation MGS > C > S > K > D, meaning that words in an MGS (for instance) are learned younger and more concrete, frequent, and imageable than words in the Core, the Satellites, the Kernel, or the whole Dictionary minus the Kernel (D – K).

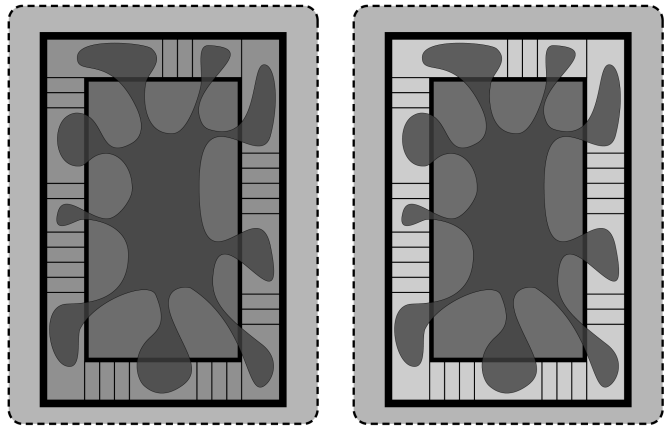

**Fig. 2. Moving inward toward the Core, words are more concrete, more frequent, and learned younger**. *Left:* Based on data from the MRC Psycholinguistic database (Wilson 1987), the general pattern observed is that compared to the words in the remaining 90% of the dictionary (LDOCE, Proctor 1981), the words in the Kernel tend to be learned at a significantly younger age, more concrete, more imageable, and more frequent, both orally and in writing. The darkness level in the figure indicates the size and direction of the difference, which is about the same for all five variables: MGS > C > S > K > D. (Fig. 1 has arrows pointing to each of these structures). *Right:* All 5 variables are intercorrelated, however, so when the comparison is done with a multiple regression analysis that measures each variable's contribution independently, the pattern is similar, but the difference in imageability and oral frequency becomes insignificant and a significant reversal in one of the variables (concreteness) emerges: Those Satellite words that are uncorrelated with age of acquisition or frequency tend to be significantly more *abstract* than the rest of the dictionary. This figure illustrates the pattern schematically; the quantitative data are shown in Fig. 3. Only one MGS is shown; the pattern is similar for all three MGSs tested.

The five psycholinguistic variables are all highly inter-correlated, however, so in order to test their effects independently of one another, we performed a step-wise multiple regression analysis, introducing one variable at a time according to its strength in accounting for the variance (Fig. 3). In the comparison of the Kernel vs. the rest of the dictionary with all 5 variables, 83% of the variance was accounted for, but two of the variables (imageability and oral frequency) no longer made a significant contribution -- nor did they do so in any of the other stepwise regressions; so we drop them from our analysis and interpretation. Age made the biggest contribution, in the same direction as in the ANOVAs, the Kernel words being (acquired) younger than the rest of the dictionary. The second strongest variable was written frequency, likewise in the same direction as the ANOVAs; and the third variable was concreteness. The independent predictive contribution of all three variables was significant (p < .001). However, the direction of the concreteness variable was reversed: For the component of the variance *not* correlated with age or frequency, the Kernel words turn out to be *more abstract* than the rest of the dictionary (Fig. 3a).

This significant reversal in the Satellite layer was the only one observed in the stepwise regressions. For the Core versus the rest of the dictionary (see Fig. 3b), the directions of the independent effects in the stepwise regression were the same as they were in the ANOVAs for age, concreteness and written frequency. The same was true for the MGS versus the rest of the dictionary, except that the effect of concreteness was very weak (see Fig. 3c).

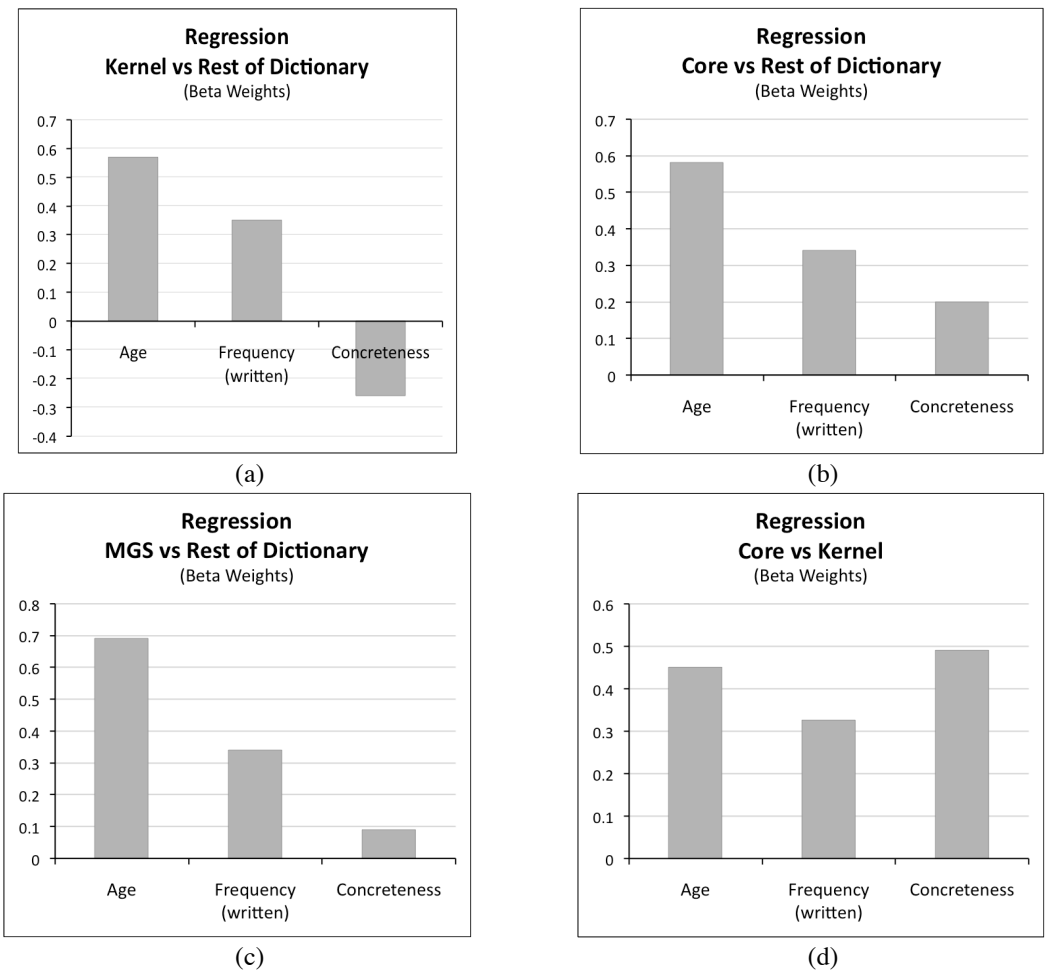

**Fig. 3. Independent regression analysis of psycholinguistic differences**. Stepwise linear regression reveals that the words in the Kernel (a), Core (b) and MGS (c) of the dictionary are learned significantly younger and are more frequent in writing than words in the rest of the dictionary. Pairwise ANOVAs show that the Kernel, Core and MGS words are also more concrete than the rest of the Dictionary, but the regression analysis shows a reversal for concreteness in the Kernel (a). Since the Core is more concrete than the Kernel (d), the likely cause of the reversal for concreteness is that the Satellite layer of SCCs that is located between the Core and the Kernel is more abstract (see text for discussion).

The regression results for the Core versus the rest of the Kernel were also in the same direction as the ANOVAs, but in this comparison, the biggest effect of the three variables was for concreteness. In all the other comparisons the biggest effect had been that of age. The regressions comparing MGSs to the rest of the Kernel were inconclusive.

We accordingly conclude that in the Satellite layer of the Kernel, the words whose acquisition is uncorrelated with age or frequency are more abstract. The Core words may be the more concrete and frequent words that must be learned early, whereas the Satellite words that are not learned early may be more abstract because they are the kinds of words needed, in addition to Core words, in order to form MGSs that can generate the meanings of all other words -- and these Satellite words continue to grow throughout the life cycle.

## 4 Discussion

Our findings suggest that in addition to the overall tendency for words to be younger, more concrete and more frequent as one moves from the outer 90% of the dictionary to the Kernel to the Core to the MGSs, something importantly different may be happening at the Satellite layer, which, unlike the deeper layers (Core and MGS) is more abstract than the rest of the dictionary, rather than more concrete, like the rest of the Kernel (for words not learned at any particular age). It is almost certain that the Core is the most concrete of all, and that the MGSs are somewhat less concrete because, besides containing Core words, they also contain some Satellite words, which are more abstract.

These results have implications for the understanding of symbol grounding and the learning and mental representation of meanings. In order for language users to learn and understand the meaning of words from verbal definitions, they have to have the vocabulary to understand the words in the definitions, or at least to understand the definitions of the words in the definitions, and so on. They need an already grounded set of word meanings sufficient to carry them on, verbally, to the meaning of any other word in the language, if they are to learn its meaning through words alone. A grounding set clearly has to be acquired before it is true that all possible further words can be acquired verbally; hence it makes sense if the grounding set needs to be acquired earlier. It also makes sense that the words in the grounding set are more frequently used, possibly because they are used more often to define other words, especially initially (and perhaps even moreso when used formally, in writing, rather than orally).

That the grounding words are more concrete is also to be expected, because word meanings that do not come from verbal definitions have to be acquired by nonverbal means, and those nonverbal means are likely to be the learning of categories through direct sensorimotor experience: learning what to do and not do with what kind of thing. It is easy, then, to associate a category that one has already learned nonverbally with the (arbitrary) name that a language community agrees to call it (Blondin-Massé et al. 2013). The words denoting sensorimotor categories are hence likely to be more concrete (although they may not necessarily be visually imageable, as there are other concrete sensorimotor modalities too, such as hearing, touch and movement).

Categorization itself, however, is by its nature also *abstraction*: To abstract is to single out some properties of a thing, and ignore others. The way we learn what kinds of things there are, and what to do and not do with them, is not by simply memorizing raw sensorimotor experiences by rote. We learn through trial and error sensorimotor interactions to abstract the invariant properties of sensorimotor experiences that determine whether or not an instance is a member of a category, and we learn to ignore the rest of the sensorimotor variation as irrelevant. The process of abstraction in the service of categorization leads in turn to higher-order categories, which are hence more likely to be verbal ones rather than purely sensorimotor ones. For example, we can have a preverbal category for "bananas" and "apples," based on the differing sensorimotor actions needed to eat them; but the higher-order category "fruit" is not as evident at a nonverbal level, being more abstract. It is also likely that having abstracted the sensorimotor properties that distinguish the members and nonmembers of a concrete category nonverbally, we will not just give the members of the category a name, but we may go on to abstract and name their properties (yellow, red, round, elongated) too. It may be that some of these higher-order category names for more abstract categories are as essential in forming a grounding set as the more concrete categories and their names.

Finally, the lexicon of the language – our repertoire of categories – is open-ended and always growing. To understand the grounding of meaning it will be necessary not only to look at the growth across time of the vocabulary (both receptive and productive) of the child, adolescent and adult, but also the growth across time of the vocabulary of the language itself (diachronic linguistics), to understand which words are necessary, and when, in order to have the full lexical power to define all the rest (Levary et al. 2012). We have discussed Minimal Grounding Sets (MGSs), and it is clear that there are potentially very many of these; but it is not clear that anyone uses just one MGS, or could actually manage to learn everything verbally knowing just an MGS. Perhaps we need some redundancy in our Grounding Sets. The Kernel, after all, is only twice as big as an MGS. And perhaps we don't even need a full Grounding Set in order to get by, verbally; maybe we can manage with gaps. Certainly the child must, at least initially. Nor is it clear -- even if we have full mastery of enough MGSs or a Kernel -- that the best way to learn the meaning of all subsequent words is from verbal definitions alone. Although language may well have evolved in order to make something like that possible in principle -- acquiring new categories purely by verbal "telling," without sensorimotor "showing" (Blondin-Massé et al. 2013) -- in practice the learning of new word meanings may still draw on some hybrid show-and-telling.

**Limitations.** Many approximations and simplifications have to be taken into account in interpreting these findings. We are treating a definition as an unordered string of words, excluding functional (stop) words and not making use of any syntactic structure. Many words have multiple meanings, and we are using only the first meaning of each word. The MRC psycholinguistic database only provides data for about a quarter of all the words in the dictionary. The problem of extracting MGSs is NP-hard. In the special case of dictionary graphs -- and thanks also to the empirical fact that the Core turns out to be so big, and surrounded by small Satellites -- we have been able, using the algorithm of Lin & Jou (2000) and techniques from integer linear programming (e.g., Nemhauser & Wolsey 1999), to extract a number of MGSs for the small dictionary whose results we are reporting here (LDOCE). This analysis needs to be extended to a larger number of independent MGSs, to other, bigger dictionaries, such as Merriam-Webster and WordNet (Fellbaum 2010), as well as to other languages. These further studies are underway (Table 2). Note that to compute MGSs of dictionaries as large as Merriam-Webster and WordNet, one will need sophisticated techniques from integer linear programming and combinatorial optimization: we will report on these in subsequent articles.

**Table 2**. For all four full dictionaries of natural languages analyzed to date, the Kernel is less than 10% (5-9%) of the dictionary as a whole, the Core (biggest SCC) and its Satellites (small SCCs) are each about half the size of the Kernel (39-61%), and each MGS (part-Core, part-Satellites) is also about half the size of the Kernel. (LDOCE: Longman Dictionary of Contemporary English; CIDE: Cambridge Dictionary of Contemporary English; MWC: Merriam-Webster Dictionary; WN: WordNet)

|  | Dictionary Name | | | |
| --- | --- | --- | --- | --- |
|  | **LDOCE** | **CIDE** | **MWC** | **WN** |
| **Whole Dictionary (D)** Number of words | 70545 | 47988 | 249739 | 132477 |
| **Kernel (K)** Word count | 4656 | 4169 | 13181 | 12015 |
| **%D** | **7%** | **9%** | **5%** | **9%** |
| **Satellites (S)** - all small SCCs Word count | 2762 | 2042 | 5028 | 5623 |
| **%D** | **4%** | **5%** | **2%** | **4%** |
| %K | 59% | 49% | 38% | 47% |
| **Core (C)** - largest SCC Wordcount | 1894 | 2127 | 8153 | 6392 |
| **%D** | **3%** | **4%** | **3%** | **5%** |
| %K | 41% | 51% | 62% | 53% |
| **MGS** - Minimal Grounding Set Word count | 2254 | 1973 | future work | future work |
| **%D** | **3%** | **4%** | | |
| %K | 48% | 47% | | |

# 5  Future Work

In order to compare the emerging hidden structure of dictionaries with the way word meaning is represented in the mind (the "mental lexicon") we have also created an online dictionary game in which the player is given a word to define; they must then define the words they used to define the word, and so on, until they have defined all the words they have used. This generates a mini-dictionary of a much more tractable size (usually less than 500 words; Figs. 4 & 5).[5]

We are currently performing the same analyses on these much smaller mini-dictionaries, to derive the Kernel, Core, Satellites and MGSs and their psycholinguistic correlates (age, concreteness, imageability, oral/written frequency), to determine whether these inner "mental" dictionaries share the hidden structure and function that we are discovering in the formal external lexicon (see Figs. 4 & 5). These mini-dictionaries will also allow us to analyze the difference in functional role among the words in the various components of the hidden structures by examining all the individual words, which is impossible with full-size dictionaries.

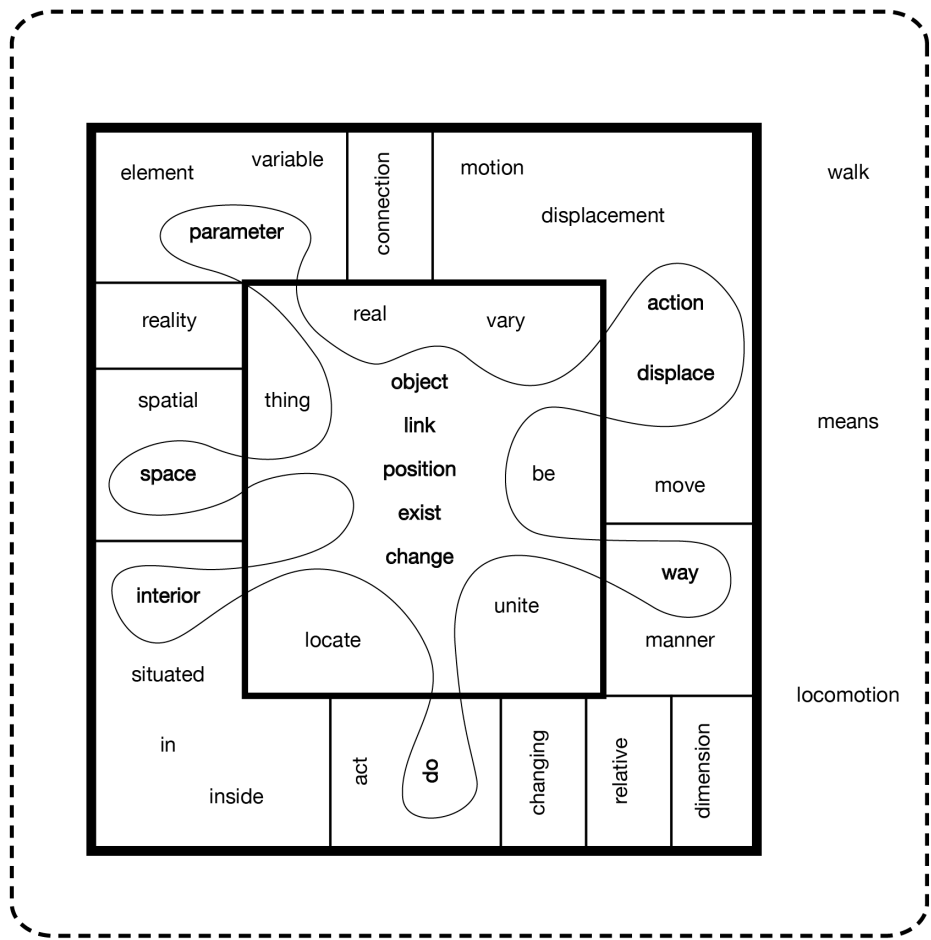

**Fig. 4. Mini-dictionary Diagram**. The diagram is the same as Fig. 1, but with real words to provide a concrete example. This 37-word mini-dictionary was generated by a player of an online dictionary game. The player is given a word and must define that word, as well as all the words used to define it, and so on, until all the words used are defined. The smallest resulting dictionary so far (37 words) is used here to illustrate the mini-dictionary's Kernel and Core plus one of its MGSs. Note that all the words words in this mini-dictionary are in the Kernel except the start word, "walk," plus "locomotion" and "means." Fig. 5 displays the graph for this mini-dictionary.

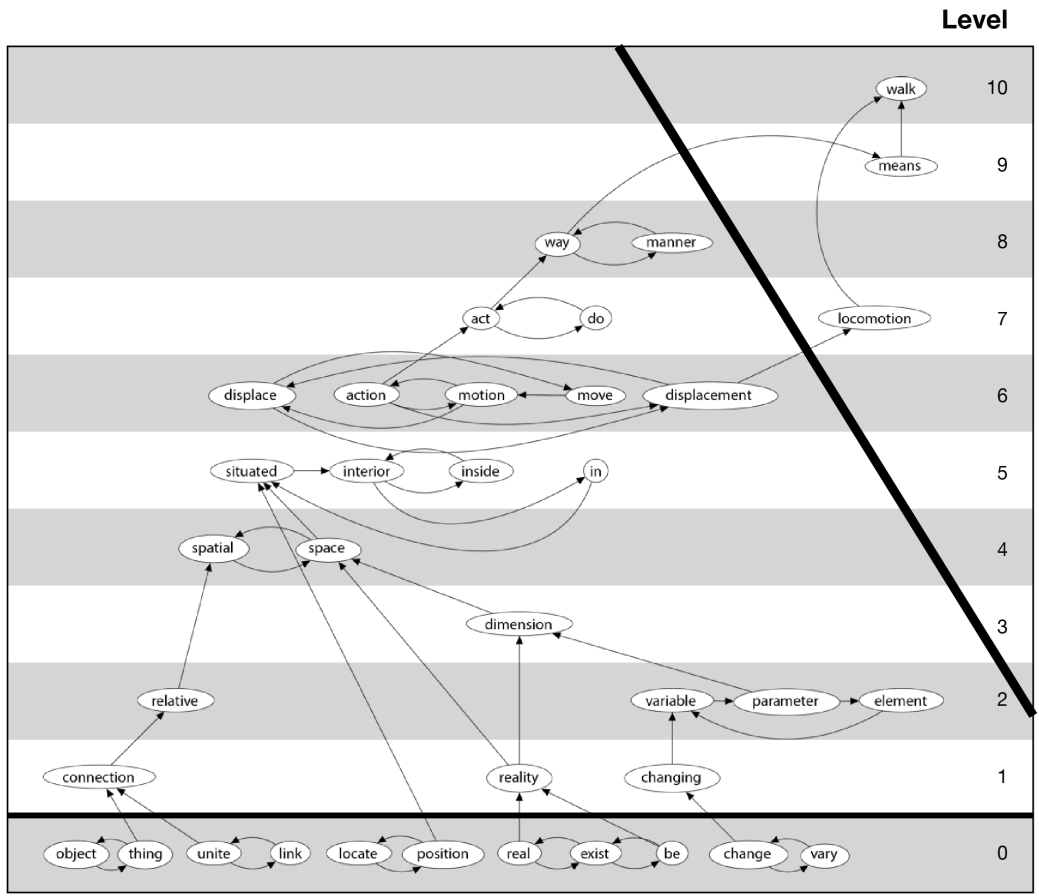

**Fig. 5. Mini-dictionary Graph**. Graph of mini-dictionary in Fig. 4, showing the definitional links. Note that in this especially tiny mini-dictionary, unlike in the full dictionaries and many of the other mini-dictionaries, the words in the Core (level 0), rather than being the single largest SCC, are the union of multiple SCCs. The oblique boldface line separates the Kernel from the (three) words in the rest of this mini-dictionary.

---

[1] Almost all the words in a dictionary (whether nouns, verbs, adjectives or adverbs) are "content" words, i.e., they are the names of categories (Harnad 2005). Categories are kinds of things, both concrete and abstract (objects, properties, actions, events, states). The only words that are not the names of categories are logical and grammatical "function" words such as *if*, *is*, *the*, *and*, *not*. Our analysis is based solely on the content words; function ("stop") words are omitted.

[2] Formally, the Core is defined as the union of all the strongly connected components (SCCs) of the Kernel that do not receive any incoming definitional links from outside themselves. (In graph-theoretical language: there is no incoming arc into the Core, i.e., there is no definitional link from a word not in the Core to a word in the Core.) It turns out to be an empirical fact about all the full-sized dictionaries we have analyzed so far, however, that their Core is itself always an SCC, and also by far the largest of the SCCs in the Kernel, the rest of which look like many small satellites surrounding one big planet (Fig. 1).

[3] In some of the mini-dictionaries generated in our online dictionary game, however, the Core is not an SCC, but a disjoint union of SCCs (Figs. 4 & 5).

[4] Most of the properties described here are empirically observed properties of Dictionary graphs, not necessary properties of directed graphs in general.

[5] The 37-word mini-dictionary in Figs. 4 & 5 is displayed because it is small enough to illustrate the hidden dictionary graph structure at a glance (and the referees asked for a real example). It was generated before we had added a new rule that a definition is not allowed to be just a synonym: In the more recent version of the game a definition has to be at least two content words (and we may eventually also rule out second-order circularity [A = B + C, B = C + notA, C = A + notB]. But it has to be borne in mind that (because of the symbol grounding problem) every dictionary is necessarily *approximate* and (at some level) *circular* (much the way all SCCs are circular). This is true whether it is a full dictionary or a game mini-dictionary generated by one player. Definitions can only convey new meanings if the mind already has enough old meanings, grounded by some means other than definition.